\documentclass[twoside,11pt]{article}

\usepackage{dnd}
\usepackage{times}
\usepackage{multicol}

\begin{document}

\title{Report from the NSF Future Directions Workshop on Automatic Evaluation of Dialog: Research Directions and Challenges}

\author{\centering
Shikib Mehri$^1$,
Jinho Choi$^2$,
Luis Fernando D'Haro$^3$,
Jan Deriu$^4$,
Maxine Eskenazi$^1$, \\
Milica Gasic$^5$,
Kallirroi Georgila$^6$,
Dilek Hakkani-Tur$^7$,
Zekang Li$^8$,
Verena Rieser$^9$,
Samira Shaikh$^{10}$,
David Traum$^6$,
Yi-Ting Yeh$^1$,
Zhou Yu$^{11}$,
Yizhe Zhang$^{12}$,
Chen Zhang$^{13}$ \\
$~$\\
    {$^1$Carnegie Mellon University}, \\
    {$^2$Emory University}, \\
    {$^3$Universidad Politécnica de Madrid}, \\ {$^4$Zurich University of Applied Sciences}, \\  {$^5$Heinrich Heine University Dusseldorf}, \\ {$^6$University of Southern California}, \\ {$^7$Amazon Alexa AI}, \\ 
    {$^8$University of Chinese Academy of Sciences}, \\ {$^9$Heriot-Watt University}, \\ {$^{10}$University of North Carolina at Charlotte}, \\ {$^{11}$Columbia University}, \\
    {$^{12}$Microsoft Research}, \\ {$^{13}$National University of Singapore} \\
}
\maketitle

\begin{abstract}%
This is a report on the \textit{NSF Future Directions Workshop on Automatic Evaluation of Dialog}. The workshop explored the current state of the art along with its limitations and suggested promising directions for future work in this important and very rapidly changing area of research.
\end{abstract}

\section{Introduction}
The \textit{NSF Future Directions Workshop on Automatic Evaluation of Dialog}\footnote{\url{http://dialrc.org/AED/}} took place virtually on July 7 - 8, 2021. This workshop was organized by Shikib Mehri, Yi-Ting Yeh and Maxine Eskenazi of Carnegie Mellon University. The participants in the workshop have extensive experience in dialog assessment. The workshop consisted of discussions and presentations which (1) explored the current state of the art in dialog evaluation, summarizing work in a field that evolves rapidly, (2) identified the limitations of existing evaluation metrics, and (3) suggested promising research directions for future work in dialog evaluation. 

This report summarizes the discussions and presentations in the workshop, with an emphasis on future research directions that funding agencies should consider going forward. The report addresses the following areas: automatic metrics in practice, assessment of evaluation metrics, human evaluation vs automatic evaluation, and the future of evaluation metrics.

The current interest in a workshop on dialog evaluation metrics began in 2016, when \citet{liu-etal-2016-evaluate} empirically demonstrated that standard automatic metrics that were commonly used to evaluate natural language generation (e.g., machine translation, summarization), such as BLEU, METEOR, and ROUGE, were insufficient and indeed misleading when used to assess state-of-the-art dialog systems~\citep{liu-etal-2016-evaluate,lowe-etal-2017-towards}. In 2018 - 2019, the interest in automatic evaluation for dialog intensified, with several papers simultaneously proposing more meaningful dialog evaluation metrics. This led to a concentrated effort to study the problem of dialog evaluation, define more meaningful automatic metrics and create datasets for the assessment of evaluation metrics. Over the past two years there has been considerable progress towards meaningful automatic evaluation metrics for dialog, with the introduction of reference-free, model-based metrics which measure multiple different dialog qualities (e.g., relevance, fluency, engagingness) and achieve moderate correlation with human judgments \citep{yeh-etal-2021-comprehensive}. The newly-proposed automatic metrics are an improvement over word-overlap metrics, yet

\begin{enumerate}
    \item they are limited in scope (they measure a limited set of dialog qualities), 
    \item they struggle to generalize, 
    \item they are not \textit{strongly} correlated with human judgment.
\end{enumerate}

As such, significant work remains to be done to improve automatic evaluation metrics for dialog. Future directions may address these three shortcomings by (1) modifying how metrics are used in practice, (2) improving upon how metrics are assessed and compared to prior work, and (3) proposing improved model architectures or training algorithms for automatic metrics. This report discusses promising directions for future research in the domain of automatic evaluation metrics for dialog.

\section{Background}
\label{sec:background}

The evaluation of open-domain dialog is challenging. Generally, when evaluating dialog systems for specific domains or constrained settings, researchers rely on targeted evaluations, such as the evaluation of dialog state tracking~\cite{williams2016dialog} or task success rate~\cite{hastie2012metrics, bordes2016learning}.
In contrast, a good evaluation metric for open-domain dialog needs go beyond these specific aspects, and must measure broader phenomena such as whether the dialog is engaging, interesting, and coherent.
Two different approaches for evaluating open-domain dialog systems are used: automatic dialog evaluation metrics and human evaluation.

Though human evaluation is generally the most reliable approach, it is an expensive and time-consuming process that requires careful experimental design, as detailed in \S\ref{sub:human_eval} and \S\ref{sub:problems}. Human evaluation may also suffer from inconsistency \citep{walker:2007} and a lack of reproducibility, which is a consequence of both the annotators' subjectivity and ambiguity in the task instructions. To mitigate the difficulty and the inconsistency of human evaluation, it is imperative that the dialog research community creates meaningful automatic evaluation metrics.

Many natural language generation tasks, such as machine translation, frequently rely on automatic metrics to evaluate systems: BLEU \cite{papineni-etal-2002-bleu}, NIST \cite{hlt-2002-nist}, METEOR \cite{banerjee-lavie-2005-meteor}, or coreference resolution, a core relation extraction task using three benchmarking metrics: MUC \cite{vilain-etal-1995-model}, B$^3$ \cite{lrec-1998-b3}, and CEAF \cite{luo-2005-coreference}. However, automatic evaluation of dialog is significantly more challenging, due to the inherent one-to-many nature of dialog \cite{zhao-etal-2017-learning}. For a given dialog context, there are many valid responses and it is therefore impractical to select a finite set of ground-truth utterances to compare to. This makes existing NLG metrics, such as BLEU or METEOR, ineffective for assessing dialog \citet{liu-etal-2016-evaluate}. To address the shortcomings of existing metrics, recent work has proposed reference-free evaluation metrics for dialog, which rely on pre-trained models to measure the appropriateness of a response rather than comparing to a ground-truth utterance. Reference-free evaluation metrics are an improvement over word-overlap metrics, however they are limited in scope, struggle to generalize and are not \textit{strongly} correlated with human judgements. Throughout this report, we identify several issues with the current state of research on automatic evaluation metrics for dialog, and suggest future directions accordingly.

The following subsections provide more detail about both the human evaluation process and the recently proposed automatic dialog evaluation metrics.

\subsection{Human Evaluation}
\label{sub:human_eval}

A comprehensive analysis of the performance of dialog systems usually includes human evaluation.
While automatic metrics can effectively assess certain aspects of dialog, human annotators are better able to holistically evaluate the performance of dialog systems, for example by assessing whether a generated dialog is realistic and interesting~\cite{liu-etal-2016-evaluate, dinan2019second, adiwardana2020towards}.
Conducting human evaluation is a challenging process. Researchers must carefully consider a variety of aspects such as experimental design, number of participants, the relevant target group, task formulation, measuring scales, definition of target concepts, etc.

Trained experts usually demonstrate higher levels of agreement and better resulting sensitivity compared to annotation by crowdworkers~\cite{gasic2011online,welleck2019neural, deriu2021survey, banchs2016expert}.
However, it is harder to recruit trained experts.
\citet{li2019acute} demonstrates how changes in the phrasing of a question has an impact on the sensitivity of the study.
Similarly, \cite{novikova2018rankme} show that ranking based on relative magnitudes significantly improves annotator agreement over the commonly used rating method using Likert scales.
\citet{huynh2021survey,dusek2019Noise} show that we can improve the quality of annotation by improving task description.  \citet{huynh2021survey} also show the impact of worker payment. They also note that  tasks are more reproducible when code is made available.
Recently \citet{smith2022human} has suggested that differing data collection methods have varying levels of human agreement and statistical sensitivity. Each method has a different amount of human annotation hours and labor costs. Performing human evaluation correctly, remains a challenging and open problem.

\subsection{Automatic Evaluation Metrics for Dialog}
\label{sub:automatic_eval}
Automatic evaluation metrics for dialog can be divided into two classes: referenced and reference-free metrics.
Referenced metrics score a system reponse by comparing it to a reference human written utterance. These metrics are ineffective for dialog~\citep{liu-etal-2016-evaluate} due to the one-to-many nature of dialog~\cite{zhao-etal-2017-learning}.
While this can be mitigated by using multiple reference responses \citep{gupta2019investigating}, it is infeasible to collect a sufficiently large dataset which thoroughly covers the space of potential responses. To this end, reference-free metrics have been proposed to circumvent the one-to-many problem.
In this section, we will first introduce a few of the most popular referenced metrics, generally used as a metric for NLG tasks like machine translation. Next, we will discuss many different reference-free metrics that have been proposed to evaluate dialog.

BLEU~\cite{papineni-etal-2002-bleu}, METEOR~\cite{banerjee-lavie-2005-meteor}, and ROUGE~\cite{lin-2004-rouge} are popular rule-based metrics used to benchmark natural language generation systems. These metrics measure the word-overlap between system-generated responses and reference responses with n-gram precision and recall. 
These metrics are easy to use and have been widely adopted to evaluate dialog generation. 
However, these metrics assume the existence of ground-truth references, which is inappropriate for dialog evaluation since it is possible that completely different responses (i.e., with low word overlap) may be pragmatically relevant to the dialog context.
\citet{liu-etal-2016-evaluate} have demonstrated that such metrics are ineffective for dialog evaluation due to the one-to-many nature of dialog \citep{zhao2017learning}.

ADEM~\cite{lowe-etal-2017-towards} and RUBER~\cite{tao2018ruber} are early learning-based metrics. ADEM uses a recurrent neural network to directly predict the quality of system responses. RUBER is a hybrid model consisting of both a referenced metric and an unreferenced metric. BERT-RUBER~\cite{ghazarian-etal-2019-better} further improved the performance of RUBER using BERT~\cite{devlin-etal-2019-bert}. Based on BERT-RUBER, PONE~\cite{10.1145/3423168} uses a novel algorithm to sample negative examples during training and trains the metric on a dataset augmented by other NLG models.

Beyond RUBER, many metrics combine models measuring different dialog qualities to further improve performance.
USR~\cite{mehri-eskenazi-2020-usr} relies on language models measuring fluency, a dialog retrieval model determining the relevance of a response, and a fact-to-response selection model measuring whether the response is conditioned on knowledge information.
Similarly, USL-H~\cite{phy-etal-2020-deconstruct} combines models evaluating the sensibleness, likelihood, and understandability of the response.
HolisticEval~\cite{pang-etal-2020-towards} adopts different models for evaluating several qualities of dialog: \textit{context coherence}, \textit{language fluency}, \textit{response diversity}, and \textit{logical self-consistency}. D-score~\cite{zhang-etal-2021-dscore} adopts a single multitask model for evaluating various dialog qualities including \textit{context coherence}, \textit{language fluency}, \textit{logical self-consistency}, and \textit{semantic appropriateness}.
Deep AM-FM~\cite{zhang2021deep} measures both semantic similarity and response fluency and the PARADISE-style model of~\cite{walker2021modeling} uses both predicted user ratings and dialog length.
DialogRPT~\cite{gao2020dialogrpt} fine-tuned a GPT-2 based model~\cite{radford2019language} on Reddit human feedback data over different dimensions such as up-votes and replies.
PredictiveEngage~\cite{ghazarian2020predictive} incorporates an utterance-level engagement classifier along with the relevance of the response to assess the overall quality of the responses.
FED~\cite{mehri-eskenazi-2020-unsupervised} calculates the likelihood of manually-designed follow-up utterances to measure 18 fine-grained qualities of dialogs.

Researchers have designed a variety of training paradigms and model structures to further improve the model-based metrics.
MAUDE~\cite{sinha-etal-2020-learning} is trained with Noise Contrastive Estimation (NCE)~\cite{gutmann2010noise}.
DEB~\cite{sai-etal-2020-improving} augments training data with manually-created relevant responses and adversarial irrelevant responses and MDD-Eval~\cite{zhang2021mdd} adopts a teacher model to augment dialog data across different domains to achieve better performance across domains.
GRADE~\cite{huang-etal-2020-grade} and DynaEval~\cite{zhang-etal-2021-DynaEval} leverages a graph structure to better model the dialog. 
\citet{peyrard-etal-2021-better} illustrate the importance of using instance-level pairing of evaluation scores. 
They propose mechanisms that aggregate pairwise comparisons such as estimating the probability that a given system scores better than another system.
FlowScore ~\cite{li-etal-2021-conversations} and the use of sentence embeddings in ~\citet{rodriguez2021automatic} model dynamic information flow in the dialog history in order to evaluate the quality of a dialog.
FBD~\cite{xiang-etal-2021-assessing} computes the distribution-wise difference between system generated conversations and human-written conversations to evaluate performance.
AIH~\cite{li-etal-2021-addressing} insert questions about the facts and opinions mentioned in the dialog history in bot-bot conversations and employ human annotators or neural models to evaluate whether the responses are consistent.
\citet{adiwardana2020towards} proposes a new human evaluation metric called SSA and suggests perplexity on public domain social media conversations might be a good automatic metric to evaluate human qualities such as sensibleness and specificity.

In addition to metrics specifically designed for dialog evaluation, there are metrics used to evaluate general natural language generation tasks.
BERTScore~\cite{zhang2019bertscore} computes the F1 score by matching BERT embeddings between human references and system responses.
BLEURT~\cite{sellam-etal-2020-bleurt} pre-trains BERT on synthetic data and fine-tunes the model to predict a human score.
QuestEval~\cite{scialom2021questeval} accounts for factual consistency, relevance, and information selection of the generated responses.
\citet{liu-etal-2021-language} uses off-the-shelf language models to create augmented responses which are used as additional references to score generated text.
POSSCORE~\cite{liu2021posscore} proposes an embedding-based metric taking the influence of POS tagging into account to evaluate conversational search systems.

\citet{yeh-etal-2021-comprehensive} present a comprehensive study of recent automatic dialog evaluation metrics. They compare many metrics for turn-level, dialog-level, and system-level evaluation and analyze their performance on fine-grained dialog qualities.
This study notes that the performance of metrics largely depends on the datasets used. It also finds that many metrics struggled with longer dialog contexts.

The fifth track of the Tenth Dialogue System Technology Challenge~\cite{chen2021automatic} was organized to address two important problems in open-domain dialog systems, each corresponding to a subtask.
The first is how to design better automatic evaluation metrics to aid research and development cycles of dialog technologies. 
Nine teams participated in this first subtask, with similar findings to \citet{yeh-etal-2021-comprehensive}, for example underlining the advantage of combining multiple sub-metrics.
It also demonstrated that performance on test datasets is worse than on development datasets, which highlights the need to develop robust metrics that can generalize to unseen evaluation datasets.

\section{Automatic Metrics in Practice} 
\label{sec:automatic_metrics}

This section explores automatic evaluation metrics from a practical perspective. First we discuss the advantages and disadvantages of human evaluation vs automatic evaluation metrics. Next, we propose the creation of annotation guidelines, which are important for standardization and facilitating collaboration within the dialog research community. Finally, we consider strategies for improving the adoption of newly proposed automatically metrics.

\subsection{Tradeoffs of Human Evaluation vs Automatic Evaluations}
\label{sub:problems}

The core advantage of human evaluation is its accurate assessment of dialog, which results from a holistic understanding of natural language. Humans are able to digest the entire dialog context in order to meaningfully evaluate a response. In contrast, automatic evaluation metrics are generally far less accurate, due to the limited ability of neural models to understand natural language dialog contexts. While some tasks in NLP have very well-defined ground-truth answers, the one-to-many problem in dialog \citep{zhao2017learning} makes it impractical to have a \textit{single} (or even multiple) ground-truth response. As such, during dialog evaluation, it is imperative that the evaluator (either a human or a metric) is able to holistically understand the dialog context in order to ascertain the appropriateness of the response. As such, an advantage of human evaluation is the fact that humans are significantly better at understanding the dialog context, and therefore better able to evaluate the appropriateness of a response. 

Though, human evaluation is vastly superior to automatic evaluation due to its accuracy, it does have a number of weaknesses. First, a disadvantage of human evaluation is its lack of consistency, with \citet{walker:2007} identifying a high degree of inconsistency in human judgements of automatically generated outputs, where ratings often differ significantly ($p<0.001$) for the same utterance. Furthermore, the \textit{same} annotator may give different scores depending on their mood, which was observed and addressed by the winning team of the Alexa Prize Socialbot Challenge 3 \cite{finch-choi-2020-towards}. This lack of consistency diminishes the reproducibility of human evaluation. In the practical sense, it may be possible to reduce inconsistencies and increase inter-annotator agreement through well-designed human evaluation tasks \citep{novikova-etal-2018-rankme}. Nonetheless, automatic metrics are inherently more consistent and reproducible than human evaluation. Another key shortcoming of human evaluation is its cost, which makes it impractical as a means of evaluation during training/model development (i.e., for hyperparameter tuning, model selection, etc.). Automatic evaluation is generally much cheaper and faster than human evaluation. Third, though human evaluation is generally accurate, it is difficult to carry out correctly. Designing a good evaluation task may require screening annotators and performing quality checks, especially when domain-specific knowledge is required to accurately assess the quality of responses.

This comparison of human evaluation and automatic evaluation highlights the importance of this body of research. Automatic metrics are cheaper, faster, more consistent and more reproducible. Unfortunately, they are far less accurate, largely because of the limited ability of models to holistically understand the dialog. As such, by improving upon automatic evaluation metrics to address their disadvantages, the dialog research community can build a valuable alternative to human evaluation. 

User testing is a standard and well established procedure in any software development cycle. In terms of dialog systems research, human evaluation is still the most reliable and trustworthy means evaluation. As such, it is worth stressing that the objective of research into automatic metrics is not to \textit{replace} human evaluation, but to instead \textit{supplement} it with meaningful automatic evaluation, that is consistent, reproducible, efficient and cheap. For example, \citet{su-etal-2016-line} propose an evaluator that learns online from human feedback.  It queries for human feedback when under-confident in its own rating, and at the same time judges the trustworthiness of the human feedback.  This results in a more robust and accurate evaluation than pure human feedback. 

\subsection{Guidelines for Human Annotation}
\label{ss:dimensions}

An important practical aspect of dialog evaluation is consistency and reproducibility. This is particularly important for human annotation, both as a means of evaluation and to create datasets for the assessment of automatic metrics. As such, one of the first tasks to undertake before beginning human annotation is to clearly define the annotation scheme. Defining such an annotation scheme is far from straightforward, even for very narrow domains. While there are ISO standards for dialog annotation\footnote{Additional standards can be found at https://www.iso.org/committee/297592.html} (e.g., ISO 24617-2:2020 \cite{bunt2020iso} or ISO 24624:2016), these standard are difficult to follow and do not perfectly encompass the annotation challenges pertaining to evaluation. Furthermore, hiring professional annotators is costly and does not scale, which is why researchers resort to relying on crowdworkers for annotation. As such, some benchmarks go through several iterations of annotation improvements by different groups in academia and industry -- for example MultiWOZ \citep{budzianowski2018multiwoz}, MultiWOZ 2.1 \citep{eric-etal-2020-multiwoz} and MultiWOZ 2.2 \citep{zang2020multiwoz}. 

Unfortunately, there is no standardized annotation scheme for dialog evaluation, particularly with respect to the dimensions to be evaluated the corresponding definitions. Often, most large-scale human annotations (i.e., to produce a human annotated dataset for assessing evaluation metrics) define a \textit{new} annotation scheme. This is because:

\begin{itemize}
    \item There is no unified definition or terminology for some dimensions of evaluation. Researchers may use different terms or different instructions when evaluating the same things \citep{finch-choi-2020-towards}. For example, to evaluate whether a response is fluent and natural,~\citet{li-etal-2019-incremental} adopts the term \textit{Fluency},~\citet{qiu-etal-2019-training} uses the term \textit{Readability}, and~\citet{zhu-etal-2019-retrieval} adopts the term \textit{Grammaticality} in the human evaluation.
    \item The human annotation may focus on a specific aspect (e.g.,  toxicity, bias, coherence, hallucination, naturalness, etc.) thereby necessitating new definitions.
    \item The relevant dimensions for dialog evaluation will evolve as dialog systems continue to improve. For example, early dialog systems struggled with fluency and relevance, while modern state-of-the-art systems are generally more robust in these aspects. As such, evaluation of modern systems will likely focus more on dimensions like long-term coherence or common-sense reasoning, rather than fluency. 
    \item Each additional dimension included increases both the cost and duration of the annotation process. As such, it is likely that most human annotation experiments will drop certain dimensions.
    \item Introducing too many dimensions during annotation will increase cognitive load, and thereby make it more difficult for humans to evaluate a system. This may result in researchers dropping certain dimensions from evaluation, to ensure better performance on other dimensions.
    \item Different dimensions may be evaluated differently by different people. First of all, annotators often evaluate according to different scales. Humans may also provide different ratings depending on their cultural background, age, knowledge, experience, etc. Subjectivity in human annotation highlights the importance of thorough definitions. 
\end{itemize}

Furthermore, a particular human annotation may have vastly different goals. System performance can be assessed at a turn-level, dialog-level, or at a system-level \cite{li2019acute}. Simultaneously performing human annotation for both turn-level, dialog-level and system-level evaluation will drastically increase the cost and time. Depending on how the simultaneous annotation at different levels is implemented (i.e., is turn-level annotation separate from dialog-level annotation), the quality of the dataset may be decreased due to increased cognitive load for the annotator.

Due to the lack of a standardized annotation scheme and the variety of viable goals for a particular annotation, we emphasize the importance of two takeaways for human annotation of dialog. 

First, when providing a human-annotated dataset, it is important that researchers share the details of the annotation process. Human annotation for dialog evaluation is incredibly challenging, and as such in addition to needing careful specification and definitions for each of the dimensions to be evaluated, clear guidelines and documentation are necessary in every stage of the annotation process. This includes preliminary trial annotations, inter-annotator agreements, scope of annotation, definitions and an evaluation of annotated data. The annotation scheme should indicate the annotated dimensions (ideally using the standard name for the given dimension), type of annotations, number of annotators, characteristics of the annotators, data format, etc. to facilitate the incorporation, usage and comparison of the new dataset in existing benchmarks. Gathering metadata from annotators may be a valuable means of better understanding their thought process. For instance, annotators could indicate their age or if they have experience annotating dialogs. This information may allow researchers to isolate the subjectivity in annotations, by facilitating better modelling of the annotator's expectations.


Second, it is important for future work to define standardized annotation schemes, which are well-documented and encompass many important dimensions and various viable annotation goals. To facilitate future work in the design of standardized annotation schemes, this report identifies two sets of annotation schemes that target two different levels of evaluation granularity: turn level and dialog level. The turn-level evaluation dimensions are adapted from~\citep{finch-choi-2020-towards} and presented in Table~\ref{tab:turn-dims}. The dialog-level evaluation dimensions are adapted from~\citep{mehri-eskenazi-2020-unsupervised} and summarized in Table~\ref{tab:dialog-dims}. The respective publications include more detailed versions of the annotation schemes, particularly with respect to the dimensions. These two examples are not intended to be a proposed standardized annotation scheme for dialog evaluation, but merely an example of \textit{effective} annotation schemes which have been used by prior works. The set of dimensions which are evaluated are particularly relevant for future work. 

\begin{table*}[htbp!]
\centering\resizebox{\textwidth}{!}{
  \begin{tabular}{l||l}
  \bf Dimension & \bf Definition\\
  \hline\hline
  \tt Grammaticality & Responses are free of grammatical and semantic errors \\
  \tt Relevance & Responses are on-topic with the immediate dialog history \\
  \tt Informativeness & Responses produce unique and non-generic information that is specific to the dialog context \\
  \tt Emotional & Responses indicate an understanding of the user's current emotional state and \\
  \tt Understanding & provide an appropriate emotional reaction based on the current dialog context  \\
  \tt Engagingness & Responses are engaging to user and fulfill the particular conversational goals implied by the user \\
  \tt Consistency & Responses do not produce information that contradicts other information known about the system \\
  \tt Proactivity & Responses actively and appropriately move the conversation along different topics \\
  \tt Quality & The overall quality of and satisfaction with the responses \\
  \end{tabular}}
  \caption{A set of turn-level evaluation dimensions adapted from~\citep{finch-choi-2020-towards}}
  \label{tab:turn-dims}
\end{table*}

\begin{table*}[htbp!]
\centering\resizebox{\textwidth}{!}{
  \begin{tabular}{l||l}
  \bf Dimension & \bf Definition\\
  \hline\hline
  \tt Coherence & Throughout the dialog, is the system maintaining a good conversation flow \\
  \tt Error Recovery & Throughout the dialog, is the system able to recover from errors that it makes \\
  \tt Consistency & Throughout the dialog, is the system consistent in the information it provides \\
  \tt Diversity & Throughout the dialog, does the system provides a diverse range of responses \\
  \tt Topic Depth & Throughout the dialog, does the system discuss topics in depth \\
  \tt Likeability & Throughout the dialog, does the system display a likeable personality \\
  \tt Understanding & Throughout the dialog, does the system understand the user  \\
  \tt Informativeness & Throughout the dialog, does the system provide unique and non-generic information \\
  \tt Flexibility & Throughout the dialog, Is the system flexible and adaptable to the user and their interests. \\
  \tt Inquisitiveness & Throughout the dialog, does the system actively ask the user questions \\
  \tt Overall Impression & The overall quality of and satisfaction with the dialog \\
  \end{tabular}}
  \caption{A set of dialog-level evaluation dimensions adapted from~\citep{mehri-eskenazi-2020-unsupervised}}
  \label{tab:dialog-dims}
\end{table*}

\subsection{Adoption of Metrics}

Though there has been considerable progress in the development of automatic evaluation metrics for dialog in the past two years, very few of these metrics have been adopted by the general community. This raises the important question of how to encourage the research community to adopt newly developed automatic metrics. Generally, a key barrier to using new metrics is the difficulty of \textit{relative comparison} with prior work. In order to use a new metric, a researcher would either (1) need to reproduce prior work and evaluate it with the new metric, or (2) neglect the relative comparison with prior work. Since neither of these options are desirable, researchers often opt to use more commonplace metrics since they facilitate straightforward comparisons.

To deal with these issues and increase the adoption of newly proposed automatic evaluation metrics, there are two approaches which we highlight. First, competitions and leaderboards often facilitate the quick adoption of evaluation frameworks, for example DSTC9 Track 3  \citep{gunasekara2020overview} performed evaluation with the USR and FED metrics. As such, part of the onus of increasing adoption of metrics is on the organizers of challenges and the creators of corpora, as they are responsible for defining the baselines and evaluation metrics. Second, automatic metrics should be easy to use and well-documented. Repositories similar to the one\footnote{https://github.com/exe1023/DialEvalMetrics} proposed by \citet{yeh-etal-2021-comprehensive} are a vital resource to support quick adoption of new metrics. It should be relatively easy to identify the best performing evaluation metric for a particular task. This can be accomplished by building a leaderboard for automatic evaluation metrics in dialog, as outlined in \S 4.1.

\section{Assessment of Evaluation Metrics}

This section discusses several challenges and research directions pertaining to the assessment of automatic evaluation metrics. A large majority of the research on dialog evaluation aims to develop \textit{better} automatic metrics. Currently, evaluation metrics are assessed by measuring correlation with human quality judgements. In this section, we first discuss a potential benchmark for dialog evaluation metrics. Next, we discuss the present state of datasets for dialog evaluation and highlight future directions for dialog evaluation datasets. Then, we describe fine-grained evaluation and how it is an important future direction pertaining to the assessment of metrics. Finally, we identify potential alternatives to measuring correlation with human judgments, when assessing automatic metrics.

\subsection{Benchmark}


In the past two years, many automatic metrics have been proposed. However, due to the fast pace of research, there was a limited amount of consistent comparisons between these different metrics. \citet{yeh-etal-2021-comprehensive} performed a comprehensive assessment of many recently proposed evaluation metrics on multiple different datasets. To facilitate consistent comparisons, an important future direction is the construction of a benchmark for dialog evaluation. 

Recent benchmarks and leaderboards, such as GLUE~\cite{wang-etal-2018-glue}, SuperGLUE~\cite{wang2019superglue}, DialoGLUE~\cite{mehri2020dialoglue}, have enabled researchers to measure progress across several natural language processing tasks in a consistent manner. A benchmark like this for the assessment of evaluation metrics can similarly inspire new research, speed up advancements, and ensure consistent comparisons. Benchmarks for dialog evaluation have been proposed by SemEval~\cite{semeval-2019-joint}, EANCS~\cite{eancs-2021-evaluations}, and DSTC10 track5~\cite{chen2021automatic}, though there has yet to be a permanent benchmark for evaluation. A benchmark/leaderboard may also facilitate increased adoption of new metrics, as researchers in need of an automatic evaluation metric will be able to consult the leaderboard in order to choose their metric. As such, it would be valuable to include code and documentation with each metric submitted to the leaderboard.

The organization and management of a permanent benchmark for dialog evaluation is challenging. Shortly after the launch of previous NLP benchmarks, the performance of submissions came close to the level of non-expert humans, suggesting limited headroom for further research. While we have yet to observe a similar phenomenon over the past two years of development of metrics, a benchmark for evaluation may need to be periodically expanded to encompass both new datasets, new dimensions (e.g., fluency, relevance, coherence, etc.), and new means of assessment (i.e., beyond correlation, turn-level vs dialog-level vs system-level).

A future direction strongly suggested by the attendees of this workshop is the creation of a working group that convenes regularly, creates an initial benchmark/leaderboard and coordinates assessment of evaluation metrics, as well as future developments to the leaderboard.

\subsection{Datasets for Evaluation}

The choice of a dataset to train or assess a metric is crucial and can make a significant impact on the conclusions we derive from any given study. Different human-annotated datasets used for assessment consist of (i) different dialog domains (e.g., topics, terminology), (ii) responses generated by different systems (e.g., older and lower-quality response generation models vs newer state-of-the-art response generation models), (iii) different dimensions (e.g., fluency, relevance, coherence), (iv) different granularities of evaluation (e.g., turn-level, dialog-level, system-level). \citet{yeh-etal-2021-comprehensive} demonstrated that many recently proposed metrics did disproportionately well on datasets they evaluated on, suggesting that researchers may often inadvertently over-optimize for their target datasets. This means that many metrics are bound to perform worse on different domains, different response generation models, different dimensions and different granularities. 

Since datasets play such a big role in determining the conclusions that are drawn from a particular study, future work should collect large and sufficiently diverse datasets (along the aforementioned dimensions) for assessing evaluation metrics. Such datasets should contain many domains, many response generation models, multiple different fine-grained dimensions of evaluation and multiple different granularities of evaluation. An underlying problem in dialog evaluation is the issue of generalization, i.e., that metrics should perform well on unseen data/domains. As such, an important aspect for future dialog evaluation datasets is a concertation on generalization (i.e., a dataset that will facilitate domain transfer experiments). Furthermore, the difficulty of high-quality crowdsourcing also necessitates that a good annotation scheme is developed, with clear and precise definitions for each dialog quality being evaluated.

In addition to the construction of a sufficiently large, diverse and richly annotated dataset, the participants of this workshop strongly suggest that a reference (frequently updated website, for example) be created that lists existing datasets and their properties, as well as suggested combinations of datasets for certain types of assessments. Another desirable direction is to create a group of researchers from different sites that create the criteria mentioned above. This group would give regular updates on the website as well as at conferences attended by dialog system researchers.

\subsection{Fine-grained Evaluation}
\label{subsec:fine-grained}

Dialog quality is inherently multi-faceted \citep{walker1997paradise,mehri-eskenazi-2020-unsupervised}. There is no universal definition of \textit{good dialog} that works in vastly different domains, applications and contexts. In certain situations, for example a task-oriented dialog system for restaurant reservations, it is important to produce factual and relevant responses and less important to produce engaging utterances. In contrast, in an open-domain chit-chat dialog system, it is more imperative to produce engaging responses. The criteria for \textit{good dialog} changes depending on the application, the domain and even the individual users. \citet{mehri-eskenazi-2020-usr} demonstrate that different annotators value different fine-grained qualities (e.g., interesting, relevance, etc.) differently. As dialog systems and dialog evaluation metrics continue to progress, it imperative for the dialog research community to progress beyond subjective and ill-defined notions of dialog quality. To this end, we propose that future research directions should include the advancement of fine-grained evaluation.

While measuring an overall impression is often subjective and difficult to define universally, it is possible to decompose dialog quality into multiple fine-grained qualities. These fine-grained qualities, for example ``relevant'', ``fluent'', or ``factual''  might be defined with less subjectivity and less ambiguous definitions. We posit that a large portion of the subjectivity in dialog quality is largely a consequence of the manner in which the fine-grained qualities are combined to produce an overall impression. Decomposing dialog evaluation into these fine-grained qualities allows the same definitions and automatic metrics to be applicable in vastly different contexts. The notion of ``diversity'' or ``interestingness'' are the same in both task-oriented and open-domain dialog, however they are less important in task-oriented dialog. When using a fine-grained evaluation framework to assess a dialog system, the weighted combination of the different fine-grained qualities could be configurable based on the application and the context (e.g., certain systems do not need to be engaging).

\citet{yeh-etal-2021-comprehensive} demonstrates that model-based metrics which specifically model certain dimensions (e.g., relevance, engagingness) through specific self-supervised objectives and model architectures \citep{mehri-eskenazi-2020-usr,phy-etal-2020-deconstruct,pang-etal-2020-towards}, achieve stronger correlations with human judgements. Generally, these metrics train sub-metrics to model specific qualities and combine these sub-metrics to produce an overall impression. Relative to the aggregated score, the fine-grained sub-metrics are often better correlated with human ratings of specific dimensions. Furthermore, combining multiple sub-metrics has been shown \citep{yeh-etal-2021-comprehensive} to better correlate with the overall impression human rating. These results empirically demonstrate the efficacy of modelling fine-grained dimensions of dialog.

The future of dialog evaluation must involve fine-grained evaluation. Datasets should be constructed that specifically annotate and measure certain dialog qualities, such as those described in \S\ref{ss:dimensions}. Furthermore, metrics should be independently developed for various dialog qualities. By attempting to measure an overall impression of the dialog, evaluation metrics are modelling an ill-defined phenomenon (i.e., a certain combination of fine-grained qualities that is specific to the context of the dataset used for assessment). The participants at this workshop find it to be imperative that automatic evaluation metrics specifically consider various fine-grained qualities.

\subsection{Beyond Correlation with Human Judgements}
\label{subsec:beyond-correlation}


Currently, automatic metrics are evaluated by measuring correlation with human judgments \cite{liu-etal-2016-evaluate,novikova2017we}. That is, the outputs of the automatic metrics are compared to human ratings with Pearson/Spearman correlation. We argue that this method of assessment may not be sufficient, as correlation with human judgements does not capture the whole picture.


A central issue with human judgments is that they are not necessarily reliable. Human studies are often under-powered \cite{howcroft2021happens} and suffer from a poor conceptualisation of what they aim to measure \cite{howcroft-etal-2020-twenty}. 
\citet{clark-etal-2021-thats} show that this disagreement can stem from insufficient annotator training; while \citet{novikova2018rankme} show that annotator agreement can be improved by reducing task ambiguity, for example by using relative ranking over absolute judgement. Another source of variation could be that the disagreement is due to personal preference \cite{dethlefs2014cluster}. These issues with human judgements suggest the possibility that simply measuring the Spearman/Pearson correlation between automatic metrics and (averaged or otherwise aggregated) human judgements may lead to incorrect conclusions about the relative performance of different metrics. Though such a phenomenon has yet to be demonstrated empirically, it is imperative that future work studies whether measuring correlation with human judgements is a sufficient means of assessing metrics. 

Recently, it was shown that comparative measurements are more robust in human evaluation. That is, for the same dialog context, the human judge would be shown different responses, and they need to rank the responses~\cite{fomicheva2021eval4nlp}. As such, one conceivable alternative approach for evaluating automatic metrics is to measure how well the ranking corresponds to the human rankings of candidate responses \citep{novikova2018rankme,georgila-etal-iwsds:2019}. A ranking for a set of bots would be computed from these comparisons and the metric would have to recreate it. Another approach is to acknowledge that humans will disagree on this task, e.g. due to personal preference, and consequently develop personalised evaluation metrics based on individual preference ratings, as proposed by \cite{dethlefs2014cluster}. These approaches, despite still comparing to human judgements, are potentially more robust and therefore may provide a more reliable signal about the quality of metric.

Furthermore, if a metric does not correlate well with human judgment, it does not necessarily imply that the metric is of low quality. Low correlation could also stem from the fact that the automatic metric and the human worker are measuring different aspects of quality. As such, a more in-depth analysis should be performed. Related to this issue is the fact that measuring correlation to human judgment does not explain why the metric differs from the human score. It does not reveal the reasoning behind the scores, which in itself is a helpful feature.

The participants of this workshop suggest that future work studies the efficacy of correlation with human judgements when assessing metrics, and explores alternatives to correlation, potentially by comparing the relative rankings of humans and metrics. Furthermore, future work should perform in-depth analysis of certain metrics to identify their shortcomings and the reasons they differ from human judgements.
Another approach could be to look beyond human judgements of dialog quality, and instead focus on the impact of the interaction using extrinsic evaluation metrics (see also section~\ref{subsec:extrinsic}). For example, the quality of a tutorial dialog system is primarily assessed through the so-called ``learning gain'', i.e., students take a test before and after interacting with the system and in this way it can be reliably and objectively measured how much the system has helped them learn \cite{core-etal-iitsec:2016,georgila-etal-aamas:2019}.
Other more subjective extrinsic metrics could be how confident or interested in a topic students feel after interacting with the system \cite{core-etal-iitsec:2016,georgila-etal-aamas:2019}.
Likewise, for other types of dialog (e.g., recommender systems), domain-specific performance metrics can be used (e.g., how many times users follow the recommendations of the system and how often they use it). As with human judgements of dialog quality, we can calculate correlations with automated metrics or compare the relative rankings of automated metrics and extrinsic metrics.

\section{Future of Automatic Evaluation Metrics}

This section explores future trends in automatic metrics for dialog evaluation. We first describe the desired properties that an automated evaluation system should possess. Next, we describe each of these properties in more detail, highlighting the challenges they present. The main question posed and addressed in this section is \textit{what are promising future directions for improving the quality of automatic evaluation metrics for dialog?}

\subsection{Desired Properties}
\label{ss:desired-properties}

The participants in this workshop have curated a list of desired properties that researchers and practitioners should consider when designing future automatic dialog evaluation systems.

\begin{itemize}
    \item Strong Correlation with Human Judgements (\S\ref{subsec:good-correlation})
    \item Interpretability (\S\ref{subsec:good-interpretability})
    \item Robustness against Adversarial Attacks (\S\ref{subsec:robustness})
    \item Generalizability 
    (\S\ref{subsec:generalizable})
    \item Forward- and Backward-looking (\S\ref{subsec:forward-looking})
\end{itemize}
Additional considerations besides desired metric properties include the following:
\begin{itemize}    
    \item Extrinsic Evaluation of Social Impact (\S\ref{subsec:extrinsic})
    \item Human-model Collaboration (\S\ref{subsec:human-in-the-loop})
    \item Adaptation to Rapid Changes  
    (\S\ref{subsec:strategy-handling-change})
\end{itemize}

\subsection{Improving Correlation with Human Judgements}
\label{subsec:good-correlation}
The goal of research on automatic evaluation metrics is to develop a suitable alternative to human evaluation, which is time-consuming, costly and difficult to reproduce. If automatic metrics are to be considered sufficient as alternatives, they should be strong predictors of human judgements. Though \S\ref{subsec:beyond-correlation} argues that correlation may be a noisy means of assessment, it is still valuable to develop better correlated metrics, especially for dimensions with higher inter-annotator agreements. As such, one of the core future goals of research into automatic evaluation metrics will be to \textit{improve correlation with human judgements}. There are several potential directions for improving correlation, some of which are enumerated below:

\begin{itemize}
    \item \textbf{Improved models:} Model-based automatic evaluation metrics for dialog typically aim to model the relationship between the \textit{dialog context} and the \textit{response}. Improvements to the models can come in the form of (1) architectural improvements, (2) training algorithms, or (3) inference algorithms. Recent work has explored several promising directions of improvement including using graph-based representations of the dialog \citep{huang-etal-2020-grade,zhang-etal-2021-DynaEval}, combining multiple sub-metrics \citep{ghazarian2020predictive,mehri-eskenazi-2020-usr,phy-etal-2020-deconstruct}, better leveraging pre-trained models \citep{mehri-eskenazi-2020-unsupervised,li-etal-2021-conversations}, proposing better training objectives, and data augmentation strategies. Improving models leveraged in evaluation metrics remains a promising direction for future research.
    \item \textbf{Better training data:} \citet{yeh-etal-2021-comprehensive} identified that many recently proposed evaluation metrics perform disproportionately well on data from the dataset they were trained on. While this is not a surprising observation, it does suggest the necessity for more diverse and representative training datasets for model-based evaluation metrics. Since most evaluation metrics are trained in a self-supervised manner, this suggests that large-scale self-supervised training may be a significant future direction for automatic evaluation for dialog. In particular, improving the generalizability of automatic evaluation metrics through better training data may significantly improve correlation with human judgement. The problem of generalizability, and potential directions to tackle it, are further addressed in \S\ref{subsec:generalizable}. 
    \item \textbf{Fine-grained evaluation:} As described in \S\ref{subsec:fine-grained}, fine-grained evaluation is a promising future direction for improving the correlation of automatic metrics with human judgments. There is inherent subjectivity in many human judgments, because an `overall impression' of a response/dialog is difficult to define universally. Instead, a subjective `overall impression' can be viewed as the weighted combination of a number of objective fine-grained qualities. While some human annotators may subjectively value \textit{engagingness} more than \textit{relevance}, for example, their definitions for the fine-grained qualities are more likely to be objective. As such, by developing metrics to model individual fine-grained qualities (e.g., interesting, engaging, relevant, fluent), we can achieve better correlation with human judgement. \citet{yeh-etal-2021-comprehensive} demonstrates that many metrics \citep{mehri-eskenazi-2020-usr,phy-etal-2020-deconstruct,pang-etal-2020-towards} which specifically model certain dimensions (e.g., relevance, engagingness) and combine multiple such sub-metrics, achieve better correlations with human judgements, in terms of both overall quality and the specific dimensions.
\end{itemize}

\subsection{Interpretability}
\label{subsec:good-interpretability}
The goal of dialog evaluation is to guide researchers in developing better dialog systems. As such, interpretability is an important facet for automatic evaluation. To make automatic evaluation more interpretable, we need to develop more fine-grained and more human-centric metrics. 
Most existing automatic evaluation methods only give the ratings of the dialogs or of the responses. This is not sufficient to guide a developer as to how to improve systems. For example, when measuring consistency, evaluation methods usually decide whether responses contradict the context, however this binary output does not tell us the nature of the contradictions. \citet{li-etal-2021-addressing} propose to detect the contradictions in facts and opinions, by pointing out the specifics of the contradiction. A key future direction is to develop mechanisms of extracting more information (e.g., the reason for a particular evaluation) from automatic metrics. \citet{freitag2021experts} propose a new evaluation methodology for Machine Translation grounded in explicit error analysis, based on the Multidimensional Quality  Metrics (MQM)  framework, which provides insights on type and severity of error. The authors also show that automatic metrics based on word-embeddings have a surprisingly high correlation with MQM scores.

In general, there are strategies that can be employed to make black-box neural network metrics more interpretable and explainable. For example, incorporating attention mechanisms in a metric may potentially allow users of the metric to understand what part of the input is desirable or undesirable. One could take a step further and have a metric generate natural language feedback rather than simply providing numerical scores. 

Furthermore, fine-grained metrics convey more information and are therefore more interpretable. As such, developing fine-grained metrics is a potential means of improving interpretability. Some existing automatic evaluation methods are designed to evaluate the overall quality \citep{li-etal-2021-conversations,huang-etal-2020-grade,xiang-etal-2021-assessing} or a single dimension \citep{ghazarian2020predictive}, but this cannot represent all of the desired dimensions of a human decision. How can we design a unified automatic evaluation metric consisting of many dimensions which are relevant to humans? Such a metric must be able to generalize to vastly different dimensions without the need for architectural/training modifications. Methods like FED \citep{mehri-eskenazi-2020-unsupervised} propose to evaluate on more dimensions in a scalable manner by leveraging the implicitly learned capabilities of large-scale language models. In the interest of interpretability, future work should propose metrics which are able to easily scale to multiple dimensions, potentially through techniques like prompt tuning.

\subsection{Robustness against Gaming the Metric}
\label{subsec:robustness}

One under-researched area is the robustness of automatic evaluation metrics. There are different scenarios in which the automated metrics can be fooled. \citet{sai2019re} showed that trained metrics can be fooled by simple manipulations of the response. For instance, dropping punctuation or removing certain words does not decrease the scores produced by the automated metric. On the other hand, \citet{Deriu2022ProbingTR} showed that when an automated metric is used as a reward for reinforcement learning, the policy converges to sub optimal solutions, which are rated highly by the metric. Thus, a key future direction is to develop automatic metrics which are built to be robust against these kinds of attacks. Specifically, sub-optimal responses should not be rated highly. This future direction can be generalized to robustness against all adversarial attacks. 

Besides being robust against all adversarial attacks, there are situations where the behaviors of trained metrics should not change. For instance, the metrics should be able to handle syntactic variants of the original responses, such as paraphrases or back-translated sentences. Such variants are semantically similar to the original responses. Hence, the metric scores assigned to them should be similar to those assigned to the original sentences given the same dialog context. 

\subsection{Metric Generalization}
\label{subsec:generalizable}
Existing model-based metrics have been criticized for their inability to generalize to out-of-distribution data~\citep{yeh-etal-2021-comprehensive,zhang2021mdd,smith2022human}, i.e., they are only good at evaluating dialogs that are similar to their training data (overfit). Yet, open-domain dialog systems are becoming more versatile and exhibiting novel types of skills, such as being knowledgeable, being empathetic, and carrying out human-like daily conversations~\citep{roller-etal-2021-recipes}. To measure the quality of such conversational agents, dialog evaluation metrics are expected to conduct assessment across different types of dialog. To improve the domain generalization of the model-based metrics, future work can consider the following two perspectives: (1) From the data perspective, we can develop more sophisticated mechanisms to generate high-quality multi-domain training data. For instance,~\citet{zhang2021mdd} leverages data augmentation techniques and pseudo labeling to construct a large-scale machine-annotated multi-domain dataset of around 2 million data points, called ``MDD-Data". The dataset is generated based on four existing human-human dialog corpora and can be easily extended to new corpora. MDD-Data is shown to be beneficial to the multi-domain evaluation task. (2) From the algorithm perspective, we can design more sophisticated networks to explicitly target domain generalization. For example, in machine learning, meta-learning approaches, such as Model-Agnostic Meta-Learning (MAML)~\citep{finn2017model}, are well-known for boosting model generalizability across domains.


Besides domain generalization, ideal evaluation metrics should exhibit good generalizability along other aspects, such as evaluation dimensions and multilingulity. Besides those suggested in table~\ref{tab:turn-dims} and table~\ref{tab:dialog-dims}, there are other dimensions that are equally important, such as addressing bias and toxicity in dialog responses~\citep{dinan2021anticipating,cercas-curry-etal-2021-convabuse}. The design of evaluation metrics should take them into consideration\footnote{More discussions can be found in \S~\ref{subsec:extrinsic}}. In addition, most of the current research works focus on evaluating English dialog. An important step should be the development of dialog evaluation datasets and automatic metrics for other languages such that the evaluation metrics can generalize to different languages.

\subsection{Extrinsic Evaluation of Social Impact}\label{subsec:extrinsic}

Dialog evaluation should not only consider intrinsic evaluation properties, such as dialog quality, engagingness, etc., but should also consider its extrinsic social impact, such as issues related to safety, bias and stereotyping.
\newcite{dinan2021safetykit} introduce a `SafetyKit' which comprises a suite of tools and datasets to assess possible harms a dialog model could introduce. The paper addresses 
three  safety-sensitive phenomena where ConvAI systems can cause harm: 
 the \textsc{Instigator}, \textsc{Yea-Sayer}, and \textsc{Impostor} effects.
 In the first scenario, a dialog system generates harmful content, thereby directly instigating harm. 
One of the first and best known examples is the Microsoft AI chatbot ``Tay'', which was shut down shortly after its launch for producing offensive language \cite{miller2017taybot}.
In the second scenario, a system may respond in a harmful manner by agreeing with (or otherwise replying unsatisfactorily to) user utterances that promote negative 
content, extending an approach introduced by \cite{lee-etal-2019-exploring}.
The last scenario describes situations where users receive inappropriate expert advice, 
e.g., medical advice \cite{bickmore:safety2018}. 
This taxonomy has already inspired further work in this area \cite{sun2021safety}.

\subsection{Forward-and Backward-looking Metrics}
\label{subsec:forward-looking}
Most of the existing automatic metrics are backward-looking, which means they only look at the history of the conversation. This is reasonable for the purpose of real-time feedback. However, sometimes, we also use evaluation metrics to give delayed feedback for error analysis or model selection. Thus it is important to also look forward to the future reactions of the users to understand and assess dialog quality. \citet{eskenazi2019beyond} proposed to consider the \textit{following user utterance} for both human evaluation and automatic evaluation of task-oriented dialogs. \citet{liang2021herald} proposes to use a user's feedback, such as whether the user is engaged or not, as the assessment for dialog quality. Thus, a future direction for dialog evaluation research would be: combining the traditional system-side backward looking metrics and the user-side forward looking metrics. Specifically, we can combine such metrics through simple averaging or re-weighting based on conversational tasks.



\subsection{Human in the Loop}
\label{subsec:human-in-the-loop}

As discussed throughout this report, human evaluation is the most reliable method for assessing dialog quality but the most labor-intensive. The least labor-intensive approach would be to develop fully automated unsupervised evaluation methods, but currently these general-purpose metrics do not perform at human level, especially for complex domains.

A hybrid approach keeps humans in the loop by developing evaluation metrics using a training set labeled with human ratings (e.g., the PARADISE framework \cite{walker-etal-nle:2000}). Once dialog data and human ratings are available then models can be trained to predict human ratings on unseen data \cite{georgila-etal-iwsds:2019,georgila-etal-lrec:2020}. The training data is collected from dialog participants who interact with a dialog system (fully automated or Wizard of Oz-based) and labeled with ratings either from these participants or human observers who do not participate in dialog interaction but instead just read dialogs between human users and dialog systems and rate them on a variety of aspects. Typically, human observers’ ratings are collected through crowdsourcing which is much more cost-effective than having actual users interact with and then rate the dialog system. Another consideration is that the training corpus will be associated with a particular version of the dialog system. A new version of the system may require collecting a new training set of dialog data and ratings. This is not ideal given that dialog system development can go through several iterations until the system is ready for deployment.

It is an open research question whether the above process can be simplified and become less expensive without sacrificing the accuracy of our prediction models. Simulated data have proven to be very popular for training dialog policies through reinforcement learning \cite{schatzmann-etal-sigdial:2005, georgila-etal-interspeech:2006}, and more recently for training deep learning-based dialog models \cite{elasri-etal-interspeech:2016}. It is much more cost-effective to generate simulated data encoding new system capabilities than deploy the new version of the system and generate a new and sufficiently large training corpus with human users, although training with human users can outperform policies trained with simulated users~\cite{weisz2018efficient}. It is therefore important to determine whether using models trained on simulated data and human observers’ ratings is a viable option. In particular, we need to investigate the resulting errors in predictions of participants’ ratings when we use simulated data and observers’ ratings as training data instead of participants’ dialogs and participants’ ratings.

Recently \citet{georgila-etal-lrec:2020} performed an experiment addressing this question. Participants interacted with a smart-home assistant in a Wizard of Oz setting and rated their interactions with the system on a number of aspects, namely, intelligence, naturalness, personality, friendliness, their enjoyment, overall quality, and whether they would recommend the system to others. Then dialog observers rated these dialog on the same aspects. \citet{georgila-etal-lrec:2020} also generated simulated dialogs between dialog policies and simulated users and asked observers to rate them on the same aspects. Models for predicting human ratings were trained on the simulated dialog and the observers’ ratings, the Wizard of Oz dialogs and the observers’ ratings, and the Wizard of Oz dialogs and the participants’ ratings. These prediction models were then applied to a held-out portion of the Wizard of Oz dialogs and tested against participants’ ratings. The results of this experiment suggested that for some conversational aspects (intelligence, naturalness, overall quality) just training evaluation functions on simulated data and observers' ratings could be sufficient. This is an encouraging result which should be tested further to explore the tradeoffs involved in using simulated data versus real dialogs, and observers’ ratings versus participants’ ratings.

\subsection{Strategy for Handling Changes in the Field}
\label{subsec:strategy-handling-change}
The field of dialog system assessment is evolving very rapidly. New metrics appear every few weeks and although new datasets, since they take time to collect and curate, appear with less frequency, they still are appearing at an astoundingly fast pace considering the work involved in their creation. There are several possible avenues of action for the community going forward. First, create a website that lists all of the available metrics and datasets. While a list with a link is a beginning step, it should be accompanied by an explanation of each metric and dataset, their goals and a few references to their use. The ideal webpage would have a link to some software (for example, \citet{yeh-etal-2021-comprehensive}) that can be used to evaluate each metric. It should also have some metric or assessment of the breadth of use that could be expected from the given dataset.

\section{Conclusions}
The participants of the AED workshop have the following recommendations for future directions of research in the evaluation of dialog systems.


\begin{enumerate}
    \item[3.2] \begin{itemize}
        \item Future work that performs human annotation, either for the purposes of evaluation or collecting a dataset for assessing automatic metrics, should release their annotation scheme. The annotation scheme will generally consist of (1) a set of evaluation dimensions/qualities, (2) definitions/guidelines for the annotators, and (3) documentation about the annotation procedure (metadata about the annotators, data filtering, inter-annotator agreement).
        \item The dialog research community should define standardized annotation schemes, which are well-documented and encompass many important dimensions and annotation goals.
    \end{itemize}
    \item[3.3] \begin{itemize}
        \item To increase adoption of metrics, the organizers of challenges/competitions and the creators of corpora should identify the best automatic metric for their dataset in order to encourage future work to rely on more meaningful automatic metrics.
        \item Automatic metrics should be well-documented and easy to use, potentially released as part of an all-encompassing package, similar to the HuggingFace repository\footnote{https://huggingface.co/}.
    \end{itemize}
    \item[4.1] \begin{itemize}
        \item A benchmark/leaderboard should be released for automatic evaluation metrics for dialog. To make this benchmark be useful over a long period of time, it should evolve and adapt to the improvements of the dialog systems. A working group should be created that is tasked with to creating an initial benchmark, then convening regularly to coordinate assessment of metrics and extend the benchmark.
    \end{itemize}
    \item[4.2] \begin{itemize}
        \item Future work should create large and sufficiently diverse datasets for assessing evaluation metrics. Such datasets should include many dialog domains, many response generation models, multiple fine-grained qualities and multiple granularities of evaluation (i.e., turn-level, dialog-level, system-level).
        \item A reference should be created to list existing datasets for assessing evaluation metrics, as well as guidelines for using these datasets for assessment.
    \end{itemize}
    \item[4.3] \begin{itemize}
        \item Future research should develop metrics and collect datasets for fine-grained evaluation. The notion of the overall quality of a dialog is noisy and subjective, as such the field of automatic evaluation of dialog should move towards modelling fine-grained qualities which can be combined to form more robust and reliable evaluation metrics.

    \end{itemize}
    \item[4.4] \begin{itemize}
        \item Future work should carry out studies to assess the effectiveness of assessing metrics by measuring correlation with human judgements.
        \item Future work should explore alternatives to correlating metric scores and human ratings, for example comparing the relative rankings of humans and metrics.
        \item Future work should evaluate dialog systems and the performance of automatic metrics through extrinsic evaluations, which measure the impact of the interaction. 
    \end{itemize}
    \item[5.2] \begin{itemize}
        \item Future work should improve model-based metrics through architectural improvements, training algorithms or inference algorithms. Promising directions include graph-based representations, combining multiple sub-metrics, better leveraging pre-trained models, better training objectives and data augmentation strategies.
        \item Future work should aim to make metrics generalizable through improved training data, i.e. by leveraging large-scale open-domain dialog data.
        \item Future work should develop metrics which can model different fine-grained qualities    \end{itemize}
    \item[5.3] \begin{itemize}
        \item Future work should make metrics more interpretable, by extracting more information from automatic metrics, for example by having metrics generate explanations/justifications. 
        \item Future work should perform in-depth analyses of certain metrics to identify their shortcomings and better explain why they differ from human judgements.

    \end{itemize}
    \item[5.4] \begin{itemize}
        \item Automatic metrics should be built to be robust against adversarial attacks, wherein system developers aim to artificially get high scores from the metric.
    \end{itemize}
    \item[5.5] \begin{itemize}
        \item Future work should explore strategies of improving the generalizability and robustness of metrics, both through multi-domain datasets and through training algorithms such as MAML.
    \end{itemize}
    
     \item[5.6] \begin{itemize}
        \item Evaluation should consider extrinsic social impacts of dialog technology, e.g.\ whether models produce output that is safe, trustworthy and bias-free. 
    \end{itemize}

    \item[5.7] \begin{itemize}
        \item Future work should consider combining the traditional system-side backward looking metrics and the user-side forward looking metrics.
    \end{itemize}
    \item[5.8] \begin{itemize}
        \item Future work should explore human in the loop strategies that leverage automatic evaluation metrics. 
    \end{itemize}
    \item[5.9] \begin{itemize}
        \item Future work should create additional resources (e.g. websites) for tracking new evaluation metrics and software for evaluation and datasets that have been evaluated.
    \end{itemize}
\end{enumerate}

\section{Acknowledgement}

The NSF Future Directions Workshop on Automatic Evaluation of Dialog and this report are funded by National Science Foundation grant IIS-1934222. The opinions expressed in this paper do not necessarily reflect those of the National Science Foundation.

\bibliographystyle{plainnat}
\bibliography{anthology, bib}

\end{document}